\documentclass[runningheads]{llncs}
\usepackage[T1]{fontenc}
\usepackage{graphicx}
\usepackage{makecell}
\usepackage{multirow}
\usepackage{subfigure}

\usepackage{mathtools}

\usepackage[capitalize,noabbrev]{cleveref}
\begin{document}
\title{Ti-Patch: Tiled Physical Adversarial Patch for no-reference video quality metrics}
\titlerunning{Ti-Patch: Tiled Physical Adversarial Patch}
%
\author{Victoria Leonenkova\inst{1}\orcidID{0009-0009-3574-7300} \and
Ekaterina Shumitskaya\inst{2,3,1}\orcidID{0000-0002-6453-5616} \and
Anastasia Antsiferova\inst{3,2}\orcidID{0000-0002-1272-5135} \and
Dmitriy Vatolin\inst{3,2,1}\orcidID{0000-0002-8893-9340}}
\authorrunning{V. Leonenkova et al.}
%
\institute{Lomonosov Moscow State University, Leninskiye Gory, 1, Moscow, 119991, Russia \and
ISP RAS Research Center for Trusted Artificial Intelligence, Ivannikov Institute for System Programming of the RAS, Alexander Solzhenitsyn st., 25, Moscow, 109004, Russia \and
MSU Institute for Artificial Intelligence, Lomonosovskiy Prospekt, 27/1, Moscow, 119192, Russia\\
\email{\{victoria.leonenkova,ekaterina.shumitskaya,\\
aantsiferova,dmitriy\}@graphics.cs.msu.ru}}
%
%
%
\maketitle              
\begin{abstract}
Objective no-reference image- and video-quality metrics are crucial in many computer vision tasks. However, state-of-the-art no-reference metrics have become learning-based and are vulnerable to adversarial attacks. The vulnerability of quality metrics imposes restrictions on using such metrics in quality control systems and comparing objective algorithms. Also, using vulnerable metrics as a loss for deep learning model training can mislead training to worsen visual quality. Because of that, quality metrics testing for vulnerability is a task of current interest. This paper proposes a new method for testing quality metrics vulnerability in the physical space. To our knowledge, quality metrics were not previously tested for vulnerability to this attack; they were only tested in the pixel space. We applied a physical adversarial Ti-Patch – Tiled Patch – attack to quality metrics and did experiments both in pixel and physical space. We also performed experiments on the implementation of physical adversarial wallpaper. The proposed method can be used as additional quality metrics in vulnerability evaluation, complementing traditional subjective comparison and vulnerability tests in the pixel space. We made our code and adversarial videos available on GitHub: \url{https://github.com/leonenkova/Ti-Patch}.

\end{abstract}
\section{Introduction}
Video content is one of the most popular and widely consumed forms of online media. The enormous number of new videos published online by users daily requires controlling the quality of uploaded content. Objective video quality metrics are used for this task. They can be categorised into Full-Reference (FR) and No-Reference (NR). FR metrics utilise the original (reference) and distorted videos to compute a quality score. This score is often interpreted as the distance between the original and distorted video or the amount of distortions that appear on the distorted video compared to the original. Since these metrics require access to the original videos, they can not be utilised for content quality control in video streaming services where original videos may not exist. In contrast, NR metrics are well suited for this task because they only require the distorted video to compute the quality score. The score produced by an NR metric often reflects both the technical distortions and the aesthetic appeal of a video.

Today, state-of-the-art image- and video-quality metrics (IQA/VQA) are mostly learning-based \cite{antsiferova2022video}. They correlated more with subjective scores than traditional methods based on natural statistics and custom feature extraction. However, prior works \cite{zhang2022perceptual}\cite{Shumitskaya_2022_BMVC}\cite{DBLP:conf/iclr/ShumitskayaAV23}\cite{SHUMITSKAYA2024103913} have shown that such metrics are vulnerable to adversarial attacks. An adversarial attack on an image/video quality metric is a transformation of an image/video to increase its quality score by some objective quality metric. Attacks on image and video data are closely related since video attacks are typically run per frame. Per-frame attacks on videos are critical for quality metrics. This is because the overall metric score of the entire video depends on the metric score of each frame.
Consequently, it is necessary to alter each frame or the majority of frames to impact the metric score of the resulting adversarial video significantly. Since deep learning models allow access to the model gradient and IQA metrics used to access video quality are commonly disclosed, an attacker can easily compute an adversarial perturbation and increase the quality metric score. An attack on the image- or video-quality metric is profitable for several real-life scenarios: 
\begin{itemize}
    \item cheating in public benchmarks, where increasing the metric’s quality is easier than developing a well-performing method
    \item manipulating the results of web search because metrics are used as one of the objectives to rank images and videos
    \item increasing the size of transcoded videos stored in the cloud, where the target bitrate may depend on source video quality by some metric.
\end{itemize}

Today, explainable \cite{kalyakulina2023explainable}\cite{panfilova2024applying} and robust \cite{kettunen2019lpips}\cite{li2023sok} models are gaining popularity in many fields. Our work aims to contribute to robust models’ development in the image- and video-quality assessment task. Prior works \cite{zhang2022perceptual}\cite{Shumitskaya_2022_BMVC}\cite{DBLP:conf/iclr/ShumitskayaAV23}\cite{SHUMITSKAYA2024103913} dedicated to the vulnerability analysis of video quality metrics operated exclusively in the pixel space of videos. In this paper, we first attempt to attack NR quality metrics in the physical world. We analyse the possibility of constructing physical adversarial patches that increase the quality scores of videos captured by cameras. This attack setting enables analysis of previously unexplored vulnerability properties of learning-based NR metrics. 

Attacks on video-quality metrics in the physical world may appear in streaming services: the user can place an adversarial object in front of the camera to unfairly increase the quality score of a video, leading to manipulating the ranking of search results or the size of the transcoded video. For example, putting an adversarial wallpaper, as we further show in our paper, makes the quality of the captured video extremely high (or low, depending on the optimisation objective of the attack). Such a video may be considered of a superior quality to other videos of the service, or if its transcoded version bitrate depends on the quality of the original video, an attacker may harm the streaming service’s cloud storage and increase streaming costs. 

The main contributions of the paper can be summarised as follows:
\begin{enumerate}
    \item We proposed a new methodology and made the first attempt to attack NR quality metrics in the physical world.
    \item We proposed a tiling technique that boosted the efficiency of physical adversarial attacks.
    \item We conducted extensive experiments in both pixel and physical space and unveiled the vulnerability of the PaQ-2-PiQ \cite{ying2020patches} NR metric.
    \item We implemented a physical adversarial wallpaper and showed that it works as a real-world attack on PaQ-2-PiQ \cite{ying2020patches} quality metric.
\end{enumerate}

\section{Related Work}
Adversarial attacks on deep learning models aimed to generate an image with a small perturbation that is invisible to the human eye. In this image, the target model gives unexpected behaviour \cite{szegedy2013intriguing}. To ensure the invisibility of perturbation, adversarial attacks usually minimise $L_p$ norm ($L_0$, $L_2$, $L_{\infty}$) between adversarial and original images:

\begin{equation}
\begin{array}{l}
L_p(x, x^{adv})
=  \sqrt[^p]{\displaystyle\sum_{i=1}^{H} \displaystyle\sum_{j=1}^{W}(x_{ij} - x^{adv}_{ij})^p}, 
\end{array}
\end{equation}

where $x$ is the original image, $x^{adv}$ is the adversarial image, $H$ and $W$ are image dimensions.

Adversarial attacks that minimise $L_2$ and $L_\infty$ norms \cite{szegedy2013intriguing}\cite{goodfellow2014explaining}\cite{madry2017towards} generate perturbations of small amplitude for all pixels in the image. Therefore, transferring such attacks from pixel to physical space is challenging, as most cameras are too small to capture these perturbations. On the other hand, $L_0$-norm-restricted attacks aim to change a minimal amount of pixels with unrestricted amplitude. For example, the one-pixel attack \cite{su2019one} is an attack which changes only one or several pixels to fool the target model. Since the amplitude is high enough to be captured by a camera, it is possible to transfer adversarial effects to the physical world. For this reason, in this paper, we consider adversarial attacks with a $L_0$-norm restriction on generated perturbation.

\textbf{Adversarial patch attacks.} The most popular $L_0$ attack is an adversarial patch attack. Adversarial patches were first introduced in \cite{brown2017adversarial} for attacking image classification models. This attack aimed to generate an adversarial patch that, when applied to any image at any location, would cause misclassification errors. Since then, various methods \cite{doan2022tnt}\cite{chindaudom2020adversarialqr}\cite{karmon2018lavan} for adversarial perturbation generation have been proposed. Doan et al. \cite{doan2022tnt} showed that the space of adversarial patches and natural images intersect. Based on these findings, they proposed a method called TnT (NaTuralistic adversarial paTches) to generate less malicious-looking patches using a generative adversarial network. Chindaudom et al. \cite{chindaudom2020adversarialqr} also attempted to make the adversarial patch less suspicious to human eyes and proposed the adversarial QR patch. This patch is created using an initialisation with a QR pattern and subsequent training to make successful attacks. Karmon et al. \cite{karmon2018lavan} proposed a LaVAN (Localised and Visible Adversarial Noise) attack that generates an adversarial patch of small size but high efficiency by searching more vulnerable areas in images. Adversarial patches were then extended to custom shapes like hats \cite{komkov2021advhat} and eyeglasses \cite{pautov2019adversarial}\cite{sharif2016accessorize}. Pautov et al. \cite{pautov2019adversarial} proposed to generate white-black patches to improve transferability from pixel to physical space. Developing adversarial patch generation methods led to more sophisticated physical attacks, such as adversarial T-shirts \cite{xu2020adversarial} and adversarial make-up \cite{yin2021adv}.

Adversarial patches were applied to test models’ vulnerability in various applications: image classification \cite{brown2017adversarial}\cite{doan2022tnt}, object detection \cite{liu2018dpatch}\cite{lee2019physical}, face recognition \cite{pautov2019adversarial}\cite{sharif2016accessorize} and others. However, to the best of our knowledge, there is no work dedicated to adversarial patch attacks on image- and video-quality metrics.  

\textbf{Attacks on image and video quality metrics.} 
Attacking image-quality metrics can be formulated as increasing or decreasing metrics’ scores for perturbed images. Our work considers only attacks that increase quality-metrics scores for two reasons. First, in real-life scenarios, it is more beneficial to increase objective quality scores than decrease them (e.g. achieving higher ratings in objective benchmarks). Second, by definition, adding small perturbations to images should not reduce the visual quality; however, the steadiness of the visual quality of adversarial images under $L_p$ norms can not be guaranteed. If we attacked the metric to decrease its scores and the perturbation reduced visual quality, then the metric was correct, and the attack was unsuccessful. In the case of targeting to increase metric scores, even if the visual quality decreases, the attack success holds. 

The problem of vulnerability of image- and video-quality metrics to adversarial attacks was first raised in prior works \cite{zhang2022perceptual}\cite{Shumitskaya_2022_BMVC}. Zhang et al. \cite{zhang2022perceptual} showed that a learning-based NR model scores can be increased by iterative gradient ascent method with restriction on FR metric’s scores (they used SSIM, LPIPS or DISTS). Shumitskaya et al. \cite{Shumitskaya_2022_BMVC}\cite{DBLP:conf/iclr/ShumitskayaAV23}\cite{SHUMITSKAYA2024103913} showed the possibility of fast attacks on NR metrics using universal adversarial perturbations and trained U-Net generators. Meftah et al. \cite{meftah2023evaluating} adapted FGSM \cite{goodfellow2014explaining}, BIM \cite{kurakin2018adversarial} and PGD \cite{madry2018towards} attacks from classification to quality assessment and showed the vulnerability of CNN-based NR quality models. Korhonen et al. \cite{korhonen2022adversarial} employed a Sobel filter to hide distortions within textured regions during an iterative attack on NR metrics. Sang et al. \cite{sang2023generation} proposed an attack on NR metrics with an adaptive attack strength search. Antsiferova et al. \cite{antsiferova2023comparing} proposed a comprehensive benchmark that compares the robustness of 15 NR models to 9 adversarial attacks. The methods discussed above are white-box since they require complete knowledge of the target model architecture and access to its gradient. Recently, black-box methods for attacking video quality metrics have also been proposed \cite{yang2024exploring}\cite{ran2024black}\cite{zhang2024vulnerabilities}. However, all existing methods generate adversarial examples exclusively in pixel space. Physical attacks were not previously applied to NR video quality metrics. This work proposes a method for creating a physical patch that attacks image- and video-quality metrics. Construction of the patch requires knowledge of the quality model; thus, it works within the white-box scenario. Despite this limitation, applying known techniques \cite{Xie_2019_CVPR}\cite{Wu_2021_CVPR}\cite{Wang_2021_ICCV} to increase the transferability, our patch can be applied to attack unknown metrics.

\section{Proposed Method}
\label{propM}

\subsection{Problem formulation}

Let $x$ denote an image represented as a matrix of pixels with dimensions $C\times W\times H$, where $H$ represents the image height, $W$ – is the image width and $C$ – is the number of channels. We define a video as a sequence of images (frames) $X$ = {$x_i$} $i=1...L$, where $L$ denotes the total number of frames within the video. Within this context, an NR video quality metric is defined as a function $M$ operating from the space of matrices with dimensions $L\times C\times W\times H$ to the set of real numbers: $M$: $[0, 1]^{(L\times C\times W\times H)}$ → $R$, where image pixels are in 0-1 range. This metric $M$ associates an input video $X$ with its corresponding quality score. The following is a formulation of an adversarial patch attack on an NR quality metric:
\begin{equation}
\underset{p}{\mathrm{argmax}} \{\frac{1}{N} \displaystyle\sum_{i=1}^{N}(M(R(x_i, p)) - M(x_i) \}, \lVert p \rVert _{0} \leq C\times D\times D ,
\end{equation}
where $ R(x_i, p) $ is the random patch application operator: 
\[R(x_i, p)_{i,j} = 
  \begin{cases}
    p_{i-x_s,j-y_s},  &  \text{if } i-x_s \geq 1; i-y_s \geq 1; i-x_s \leq D; i-y_s \leq D\\
    x_{i,j},          &  \text{otherwise } \\
  \end{cases} \\
\]
$ x_s $, $ y_s $ are randomly chosen patch location from [1, $W$-$D$], [1, $H$-$D$] respectively, $ p $ is the patch we are searching for, $x_i$ are images, $N$ is the training dataset size, $M$ is the attacked metric, $C$ is the width and height of the patch. 

The goal of the proposed attack is to generate such an adversarial perturbation $p \in R^{(C \times D \times D)}$, which, added to any image, will increase the score of the metric $M$.

\subsection{Baseline method}
The baseline patch attack is an adaptation of the method proposed by Tom B. Brown et al. \cite{brown2017adversarial} to our task of attacking the NR IQA metric. The patch generation process is described as follows. For each image $x_i$ from the training dataset $D$, we randomly set up a mask, which is a matrix of dimensions $C \times W \times H$ consisting of zeros and ones, indicating where the patch $p$ will be placed into the image $x_i$. Randomising the patch's placement is provided by its position in the image and its rotation by an angle that is a multiple of 90 degrees. This randomness increases the attack's success regardless of patch positioning within the frame across various shooting conditions. The patch is trained by iterative optimising the loss function:
\begin{equation}
    loss = 1 - \frac{M(x^p_i)}{M_{range}},
\end{equation}
where $M$ is an attacked metric, $M_{range}$ is the range of metric, $x^p_i$ is the perturbed image, $x_i$ is the original image.
The perturbed image is:
\begin{equation}
        x^p_i = M^p_i \circ P^m_i - (1-M^p_i) \circ x_i,
\label{eq:perim}
\end{equation}
where $M^p_i$ is a mask for random patch application, $P^m_i$ is the patch, $x_i$ is an image from training dataset $D$. Further sections describe our proposed improvements to the baseline approach.

\subsection{Enhancement of Transferability To Physical Space}
A patch generated by a baseline method may contain prominent high-frequency components, which force the target metric to increase its score when digitally applied to image pixels. However, these high-frequency components look differently after printing on paper and cannot be captured by most cameras. Moreover, compared to a digital patch, a physical one can be affected by different light, perspective and noise distortions. Finally, the image or video captured by a camera undergoes compression, which can further affect the adversarial patch attack efficiency. In addition to the baseline patch attack, we implemented several additional features to enhance the transferability of generated patches from pixel to physical space. 

\textbf{TV.} We added Total Variation ($TV$) \cite{mahendran2015understanding} patch minimisation to the training process to reduce high-frequency components in the generated patch. $TV$ for the patch is defined in Formula \ref{eq:tv}. $TV$ aims to make neighbouring pixel values close to each other. Patches with lower $TV$ are smoother and more robust to compression.
\begin{equation}
        TV(p) = \displaystyle\sum_{i,j}((p_{i,j} - p_{i+1,j})^2 + (p_{i,j} - p_{i,j+1})^2)^{\frac{1}{2}},
\label{eq:tv}
\end{equation}
where $p_{i,j}$ is a pixel in patch $p$ at coordinates $(i, j)$.

\textbf{NPS.} The trained patch needs to be printed to implement the physical attack. However, the printer's ability to render colour is limited so that it may reduce the efficiency of patch attacks. Non-Printability Score ($NPS$) \cite{sharif2016accessorize} is an approach that brings digital colours closer to the range of colours that printers can reproduce. $NPS$ of one pixel is defined as shown in Formula \ref{eq:nps}. Consequently, the $NPS$ of the patch is defined as the sum of the $ NPS$ of all the pixels in it. $NPS$ minimisation allows the generation of the patch with better printability. 
\begin{equation}
    NPS(p_{i,j}) = \displaystyle\prod_{t\in T}| p_{i,j} - t|,
    \label{eq:nps}
\end{equation}
where $T \subset [0,1]^3$ is a set of printable RGB triplets, $p_{i,j}$ is a pixel of the patch $p$.

\textbf{Relighting.} Videos can be shot in different locations and light conditions. In other words, we need to consider not only geometric transformations but also relighting transformations. We implemented a patch generation method with random brightness augmentation to make the patch more robust to small light changes.

\textbf{Black-white.} As we mentioned in Related Work, white-black patches can be used to improve transferability from pixel to physical space (Pautov et al. \cite{pautov2019adversarial}) since white and black colours can be printed with fewer colour distortions. We tested this assumption to attack IQA metrics. We chose the best variants of patch training and generated black-white patches using these methods.

\textbf{Without rotation.} Since an attacker can control the angle of the patch during a physical attack, we decided to simplify training and remove random rotation from it. Also, a patch trained without rotation can produce higher adversarial efficiency since it is free from randomness compared to the baseline method.

\begin{figure}[tb]
\includegraphics[width=0.98\textwidth]{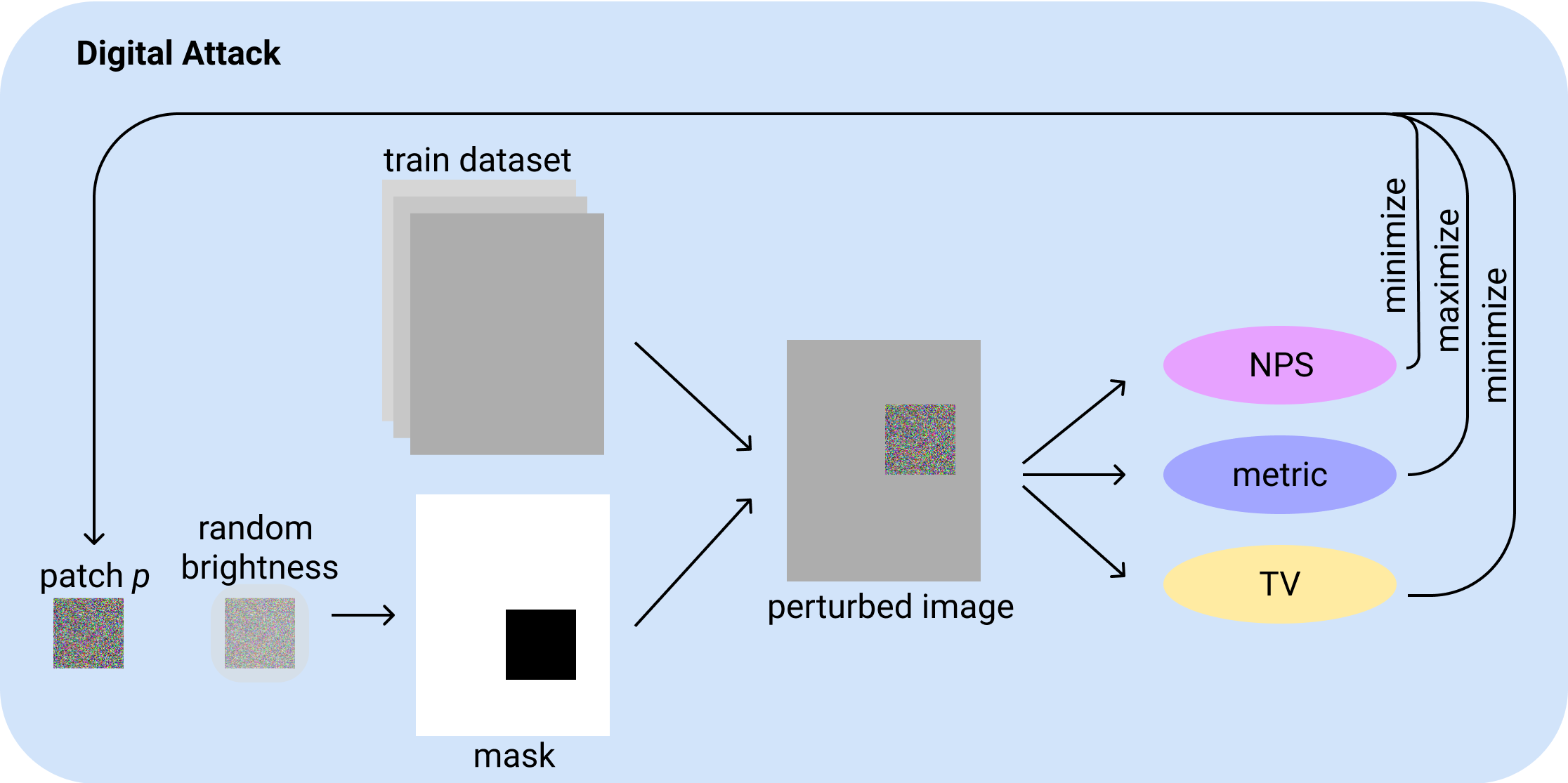}
\caption{The proposed digital attack scheme.} 
\label{digital}
\end{figure}

The general attack approach is depicted in Fig. \ref{digital}. 

\subsection{Physical Attack}
\begin{figure}[tb]
\includegraphics[width=0.98\textwidth]{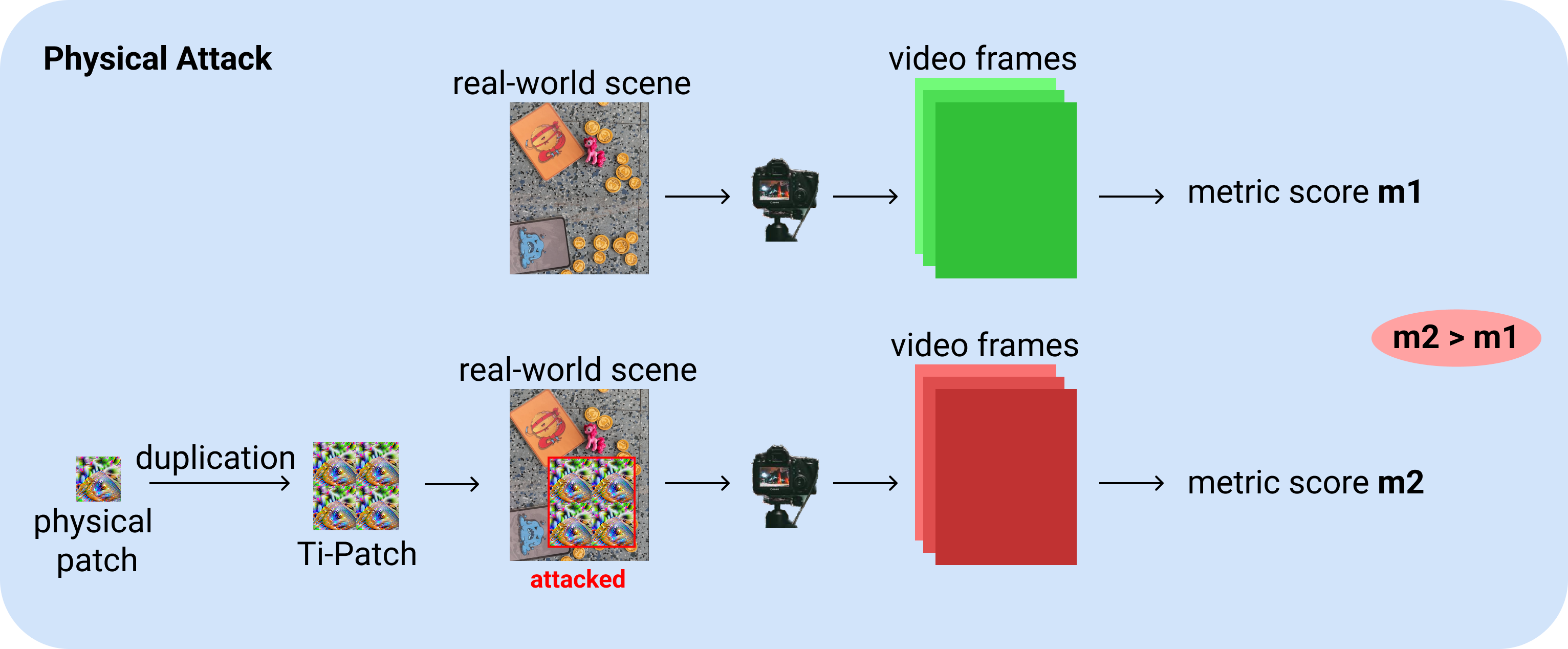}
\caption{The proposed physical attack scheme.} 
\label{physical}
\end{figure}

To fool a classifier or detector, you need one object that affects a small part of all pixels of the image. The case of attacking image quality metrics is more complicated: a metric score depends on each pixel. Therefore, attacking a metric with a small patch is a tricky task. Moreover, it makes the reproducibility of the attack in the physical world more challenging because of the high resolution of video frames. A patch of a specific size is trained to increase the metric score, so to maintain the same percentage of the patch size in a video, we suggest ``tiling'' the patch to cover the required area size in a captured video. After printing and ``tiling'', the Ti-Patch is positioned within a real-world scene, and the video is recorded. Subsequently, the metric score increases compared to the untiled patch. The trained patch can be utilised in physical attacks by printing and showing it from a phone or monitor. The physical implementation scheme is presented in Fig. \ref{physical}. 

\section{Experiments methodology}
\label{exp}

\subsection{Compared methods}
Following the ideas described in the Proposed method \ref{propM}, We generated 8 various trained patches : 
\begin{enumerate}
  \item Baseline.
  \item Baseline+TV+NPS: baseline with adding to the optimization $TV$ loss and $NPS$ loss.
  \item BaselineL: baseline with additional adjustments to patch in the form of random changing brightness.
  \item BaselineL+: baseline with $TV$ loss and $NPS$ loss and additional changing brightness.
  \item B-WBaselineL+: BaselineL+ in black-white variant.
  \item B-WBaselineWRL+: BaselineWRL+ in black-white variant.
  \item BaselineWRL+(ours): baseline without rotation with $TV$ loss and $NPS$ loss and additional changing brightness.
  \item BaselineWR(ours): baseline without random rotation patch while applying it on the image.
\end{enumerate}

\subsection{Attacked NR IQA metric}
As a target metric for our patch attacks, we used the PaQ-2-PiQ with $M_{range} = 100$ \cite{ying2020patches} as it showed high performance in MSU Benchmark \cite{NEURIPS2022_59ac9f01} and high-speed in backpropagation. 

\subsection{Datasets for pixel space experiments}
For training, we selected 1000 $256\times256$ images from the COCO dataset \cite{lin2014microsoft}. 
Validation was performed using 250 $256\times256$ images from the COCO.

Testing was done using 500 images from the COCO dataset and two additional datasets: NIPS 2017 image dataset \cite{nips-dataset} and DERF 2001 video dataset \cite{derf-dataset}. NIPS 2017 dataset consists of 1000 images with a resolution of $299\times299$. From the DERF 2001 dataset, we selected 20 FullHD videos and resized them to $640\times360$. For each video, we used the first 75 frames. 

\subsection{Evaluation metric}
We measure the success of the proposed attack for one image as a difference in metric scores before and after an attack. It is formulated as follows:
\begin{equation}
	metric\ gain = M(x^p_i) - M(x_i),
\end{equation}
where $M$ is a metric, $x^p_i$ is a perturbed image and $x_i$ is an original image.

For adversarial wallpaper setup, the metric gain is calculated between the frame with the TV set's black/white background and the current frame.

\subsection{Implementation details}
\subsubsection{Pixel space}
When evaluated, a patch was applied with a random location for all images from a dataset. Patches that were trained with random rotation were also tested with random rotation. All patches were put at the same positions in each video, while the positions and rotation parameters were chosen randomly and fixed for each video. 

\subsubsection{Physical space}
We conducted 3 types of physical experiments. In the first experiment, we applied all 8 patches without tiling. In the second experiment, we tested the best batch with tiling. Finally, we did experiments with implementing digital adversarial wallpaper. We used an iPhone 15 Pro camera, and the video resolution was $1080 \times 1920$, with 30 frames per second. Moreover, we printed patches on thick paper to avoid patch transformations.  
\begin{enumerate}
    \item \textbf{Single patches}. 
    We tested patches in 3 different locations --- two were horizontal, and one was vertical. There were 3 distances from the patch to the camera: 22, 25 and 28 centimetres. We created an adversarial garland to carry out the attack consistently, quickly and under the same conditions. The patches were glued to transparent tape at 30 centimetres from each other. Initially, the metric score of a background with tape and without patches was measured. Then, we slowly moved the tape to capture each patch in the same location (Fig.~\ref{exp} (a)). For each patch, we calculated the final gain as the maximal gain achieved when the patch was in front of the camera.
    \item \textbf{Tiled patches}. We tiled the patch to compare the metric gain of the single patch and the Ti-Patch and to test our assumption of maintaining an approximate percentage of the frame area occupied by the patch. Then, we recorded videos in the same 3 locations. At first, there was a single patch in the scene, but then the tiled version appeared. We compared the resulting gain of the two variants. The distance from the camera to the patch remained the same as in the experiments with single patches (Fig. \ref{exp} (b)). 
    \item \textbf{Adversarial wallpaper}. 
    We checked the efficiency of the generated patch on 3 long distances: 125 cm, 157 cm, and 206 cm. A patch was replicated all over the screen in different sizes. A person was sitting in front of the monitor during the experiment. By scrolling through the resulting variants and altering the distances between the TV set and the phone camera, we monitored the changes in the metric gain (Fig. \ref{exp} (c)).
\end{enumerate}

\begin{figure}[tb]
\includegraphics[width=\textwidth]{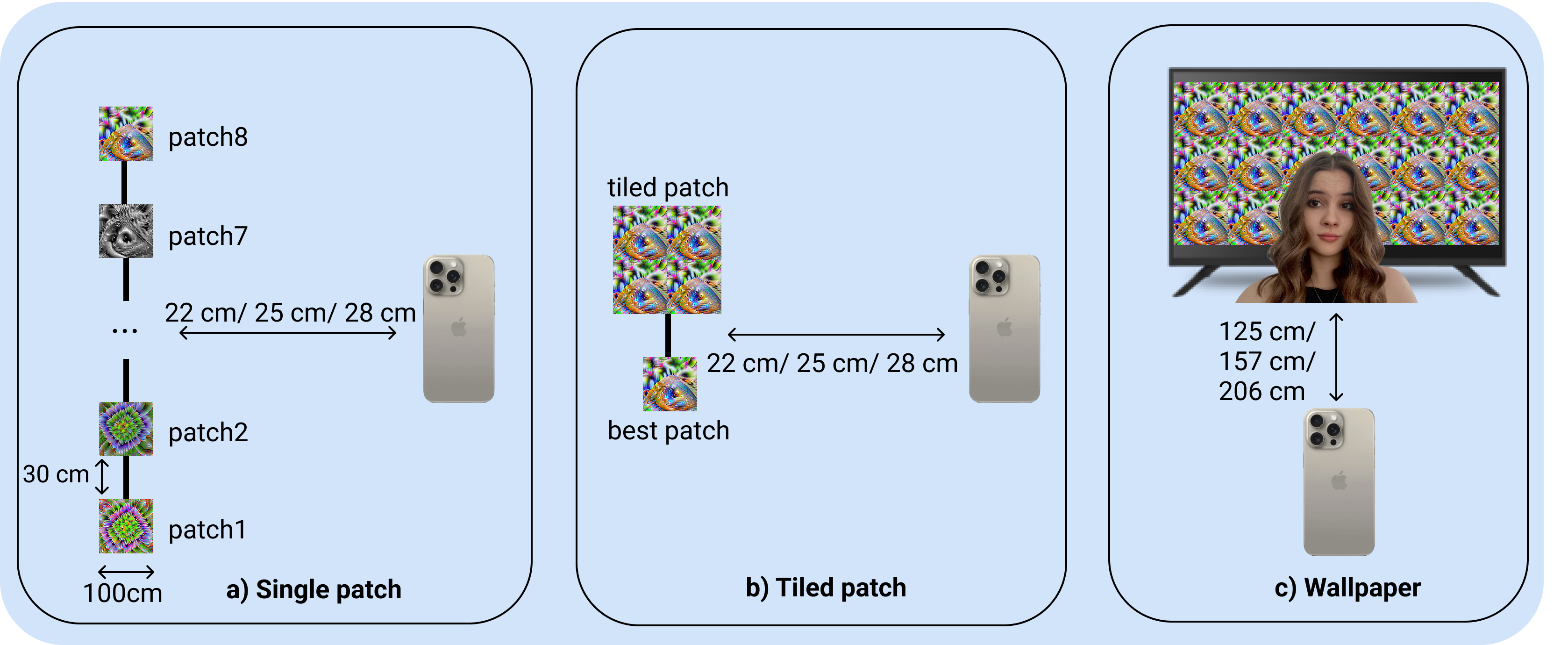}
\caption{The scheme of physical experiments.} 
\label{exp}
\end{figure}

\section{Results}
\label{res}

\begin{figure}[tb]
\includegraphics[width=0.98\textwidth]{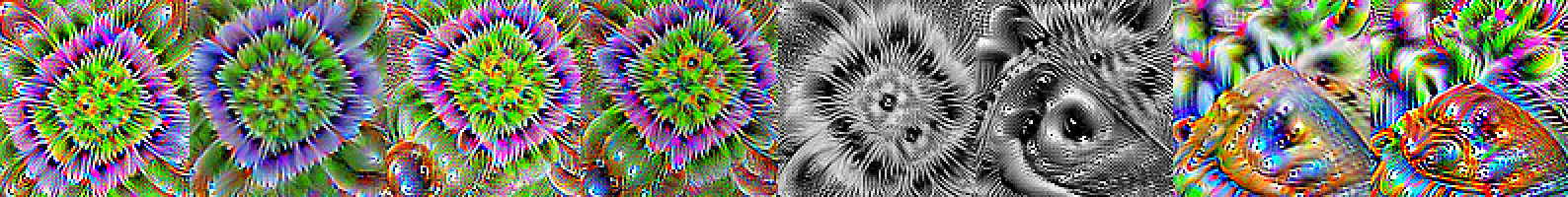}
\caption{8 generated patches to attack PaQ-2-PiQ metric. From left to right: Baseline, Baseline+TV+NPS, BaselineL, BaselineL+, B-WBaselineL+, B-WBaselineWRL+, BaselineWRL+(ours), BaselineWR(ours).} 
\label{patches}
\end{figure}

Figure \ref{patches} represents a visualisation of all generated patches.

\subsection{Pixel space}
Table \ref{tab1} shows the results for experiments conducted in pixel space. We can conclude that a patch trained on one dataset is well transferred to unseen datasets of another content and resolution. Also, we can see that white-black patches produce lower gain since image-quality metrics are probably highly sensitive to colours. However, based on results only from experiments in the pixel space, we cannot conclude which patch is better for physical space. The results of physical and pixel experiments vary due to camera characteristics, and the patch leaderboard in the physical world can change significantly. A metric gain in real-world experiments is expected to be lower than in pixel space. Our improvements to the baseline aim to increase the success of a physical attack. As the results show, in pixel space attacks, such additions reduce gain since these additions impose some restrictions on generated patches. However, as seen from the following section \ref{physExp}, these additions benefit real-world attacks.

\begin{table}[tb]
\caption{Average increase in PaQ-2-PiQ \cite{ying2020patches} metric scores after patch attack in the pixel space for eights methods and 3 datasets.}\label{tab1}
\begin{tabular}{cccccc}
\hline
& \multicolumn{2}{c}{COCO \cite{lin2014microsoft}} & NIPS \cite{nips-dataset} & DERF \cite{derf-dataset}\\
 & \makecell{train\\1,000 images} & \makecell{test\\500 images} & 1000 images & 12 videos\\
\hline
Baseline \cite{brown2017adversarial} & 17.44 $\pm$ 0.25 & 16.07 $\pm$ 0.35 & 15.23 $\pm$ 0.20 & 16.49 $\pm$ 0.26 \\
Baseline+TV+NPS & 15.44 $\pm$ 0.24 & 15.50 $\pm$ 0.31 & 13.50 $\pm$ 0.20 & 14.56 $\pm$ 0.22 \\
BaselineL & 16.36 $\pm$ 0.25 & 16.46 $\pm$ 0.35 & 14.25 $\pm$ 0.19 & 15.43 $\pm$ 0.26 \\
BaselineL+ & 16.39 $\pm$ 0.25 & 16.39 $\pm$ 0.34 & 14.22 $\pm$ 0.19 & 15.55 $\pm$ 0.25 \\
B-WBaselineL+ & 9.88 $\pm$ 0.22 & 9.88 $\pm$ 0.30 & 8.55 $\pm$ 0.16 & 10.03 $\pm$ 0.22 \\
B-WBaselineWRL+ & 18.81 $\pm$ 0.24 & 18.93 $\pm$ 0.30 & 17.23 $\pm$ 0.17 & 18.09 $\pm$ 0.23 \\
\makecell{BaselineWRL+ (ours)} & \underline{29.11 $\pm$ 0.29} &  \underline{28.85 $\pm$ 0.37} & \underline{26.30 $\pm$ 0.20} & \underline{27.47 $\pm$ 0.27} \\
BaselineWR (ours) & \textbf{30.18 $\pm$ 0.28} & \textbf{30.19 $\pm$ 0.39} & \textbf{27.29 $\pm$ 0.22} & \textbf{27.86 $\pm$ 0.25} \\
\hline
\end{tabular}
\end{table}

\subsection{Physical space}
\label{physExp}

\subsubsection{Single patches} 
\begin{figure}[tb]
\centering
\includegraphics[width=0.88\textwidth]{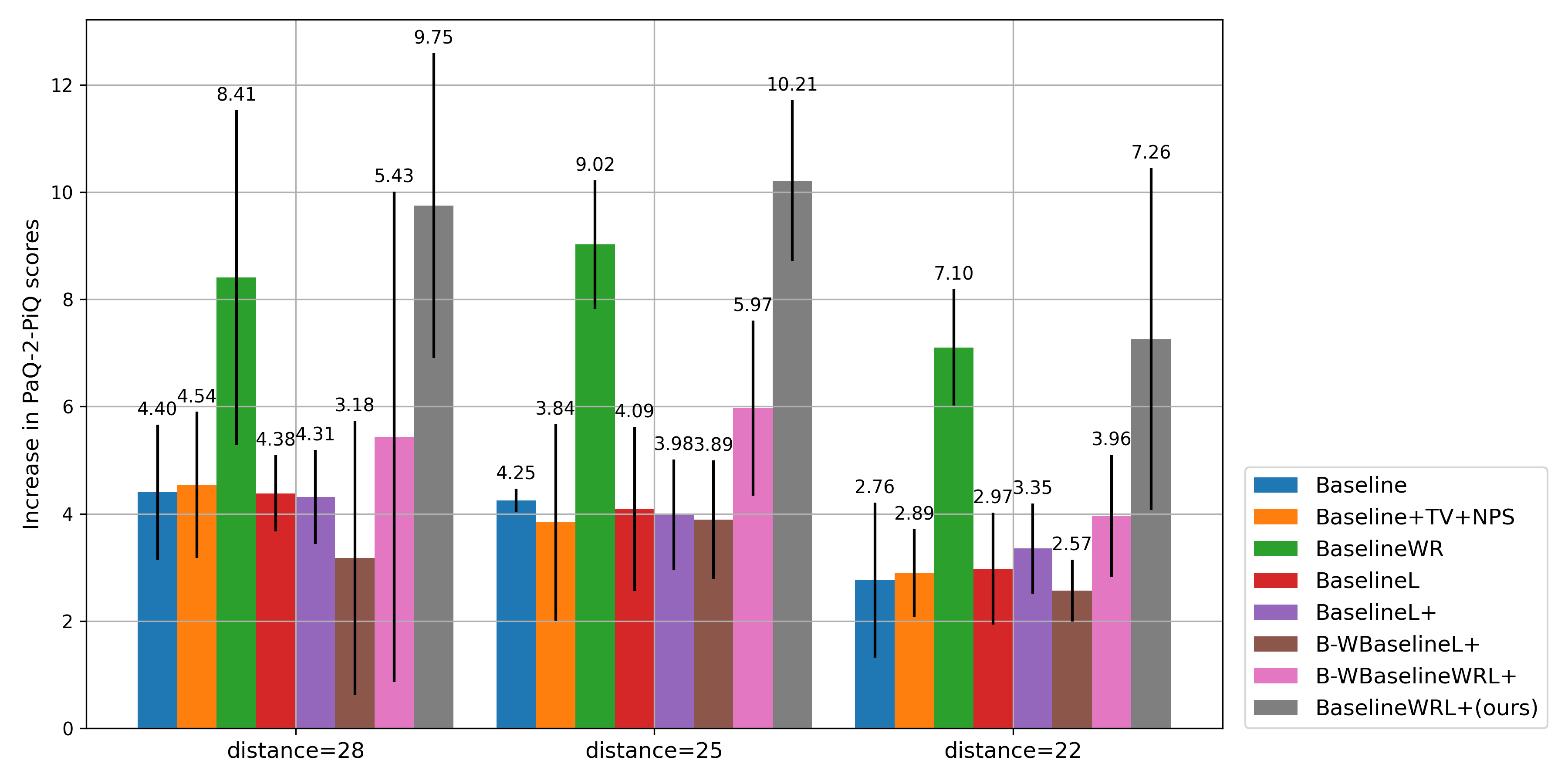}
\caption{Results of physical experiments for 8 single patches with 3 distances to the camera. Increase in metric scores averaged by 3 different locations. Vertical lines represent 95\% confidence intervals for mean scores.} 
\label{fig1}
\end{figure}

The results are presented in Fig. \ref{fig1}. We averaged the metric gain for each distance for 3 locations. From the results of this experiment, we can conclude that the patch with the highest metric gain is BaselineWRL+. Proposed patches have the position concerning the camera, in which the increase in metric scores is maximal. We call this position optimal.
On the one hand, if the camera is too close to the scene, the patch is resized to a higher size. On the other hand, if the camera is far away, the patch size is resized to a lower size. In attacks on quality metrics, patch resize affects the metric gain significantly. From the results of our experiments, we can assume the optimal position --- from 22 to 28 cm, since it is a local maxima on 25 cm distance. In this interval, uncontrolled camera resize will not significantly influence the gain. Another conclusion from the results is about black-white patches. Our expectations about better reproductivity of black-white patches in the physical world were unmet. Metric pays a lot of attention to image colours. Moreover, our additional features for physical transferability enhancement, such as $TV$, $NPS$, and relighting, have successfully contributed to higher gains in real-world experiments.

\subsubsection{Tiled patches} 
\begin{figure}[h!]
\begin{center}
\includegraphics[width=0.55\textwidth]{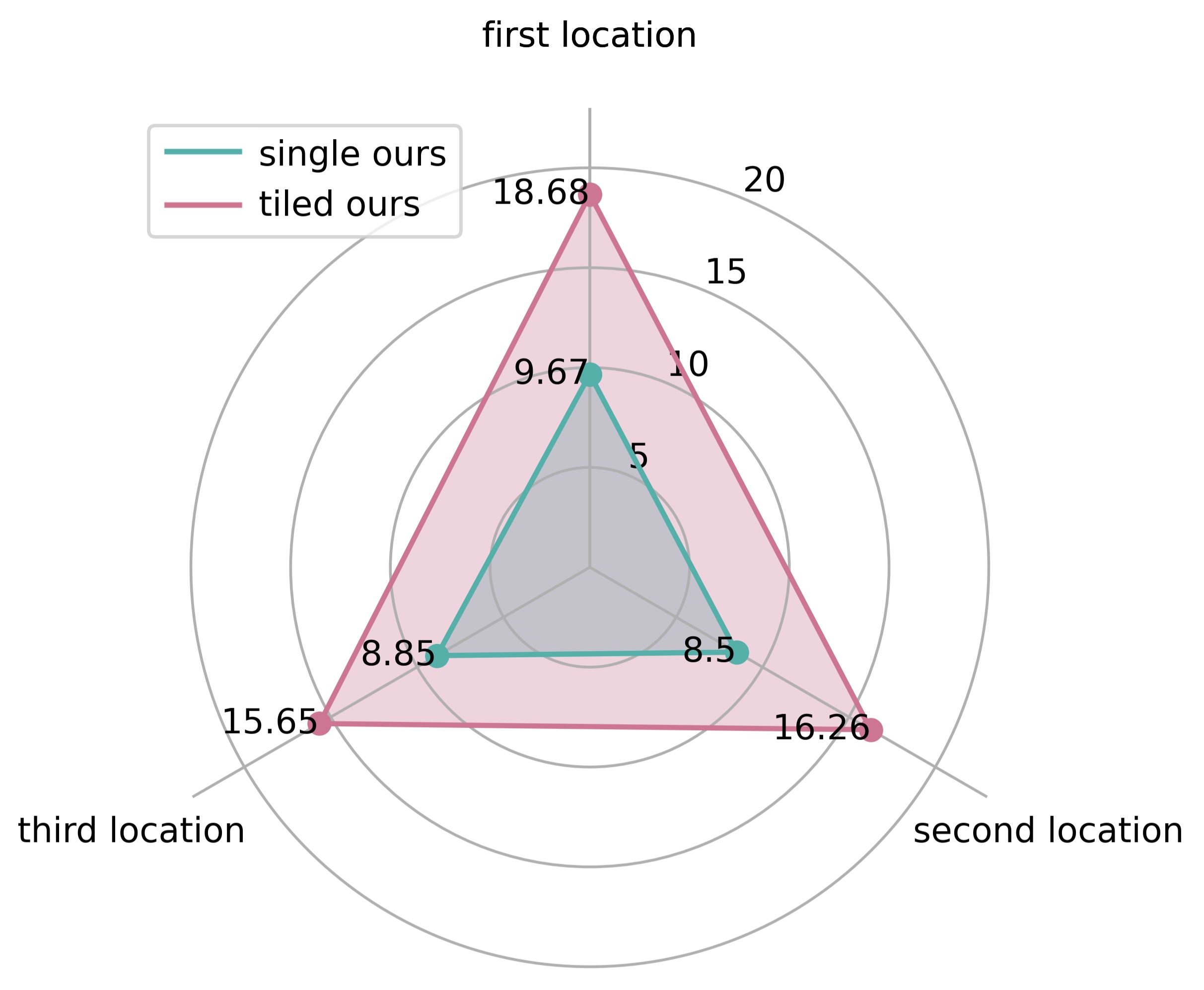}
\caption{The increase in PaQ-2-PiQ scores after physical single and tiled patch attacks on 3 locations. Results averaged by 3 distances to the camera.}
\label{rosa}
\end{center}
\end{figure}

According to the results of experiments with single patches, we chose the best patch --- BaselineWRL+(ours) and did the tiling experiment with it. The results can be seen in Fig. \ref{rosa}. There, we averaged the metric gain for each location over 3 distances. As expected, the tiled version showed a higher metric gain than the single patch. 

\subsubsection{Adversarial wallpaper} 
\begin{figure}[tb]
\includegraphics[width=0.98\textwidth]{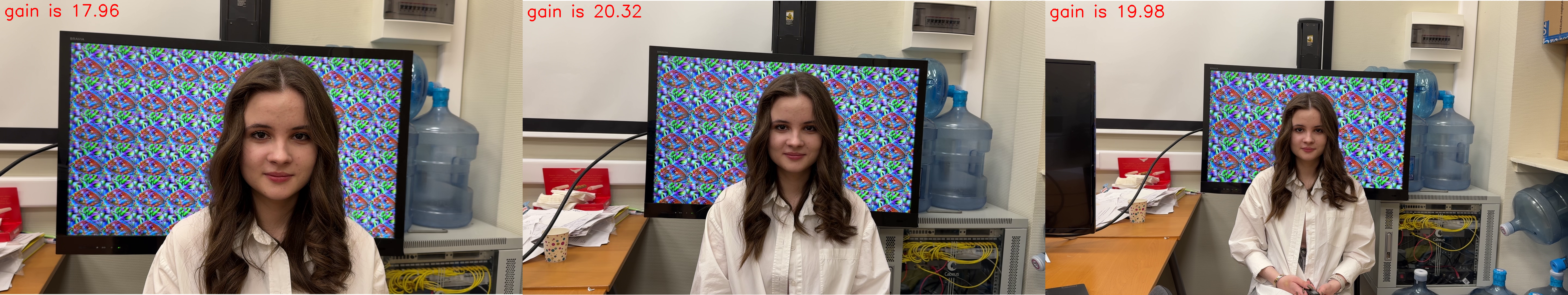}
\caption{Examples of adversarial wallpaper experiments.} 
\label{wall}
\end{figure}

The best patch, BaselineWRL+(ours), was chosen again for this experiment. The resulting frames are presented in Fig. \ref{wall}. Our patch achieved a metric gain of 18-20 on each distance, the highest among all physical settings. 

\section{Conclusion}
This paper demonstrates the applicability of physical patch attacks on the no-reference image- and video-quality metrics. According to our findings, quality metrics can be attacked not only in pixel space but in the physical world, too. We conducted extensive experiments with 8 patches in 3 locations and 3 various distances. Our proposed improvements to the baseline solution \cite{brown2017adversarial} BaselineWRL+ achieved the 20 points gain, which is 20\% of metric range. It is an excellent result for a small attacking metrics physical patch with the size of 100$\times$100 pixels.
We additionally showed that our approach works in distinct real-world experiments such as tiling and digital wallpapers.   

\textbf{Future Work}.
According to other research \cite{antsiferova2023comparing}, other metrics are easily attacked to increase its score. We plan to test the proposed method on other NR image- and video-quality metrics. Moreover, we want to create naturalistic adversarial patches, like \cite{doan2022tnt}. Furthermore, we plan to transfer our method on other regression models, such as age detectors.
%
%

\begin{credits}
\subsubsection{\ackname} The authors would like to thank the video group of MSU Graphics and Media laboratory, especially Aleksandr Kostychev for the assistance in filming the physical experiments. This study was supported by Russian Science Foundation under grant 24-21-00172, https://rscf.ru/en/project/24-21-00172/.

\subsubsection{\discintname}
The authors have no competing interests to declare that are
relevant to the content of this article.
\end{credits}
%
%
%
\bibliographystyle{splncs04}
\bibliography{egbib}

\end{document}